\documentclass[10pt,twocolumn,letterpaper]{article}

\usepackage{cvpr}              

\usepackage{times}
\usepackage{epsfig}
\usepackage{graphicx}
\usepackage{amsmath}
\usepackage{amssymb}
\usepackage{microtype}
\usepackage{graphicx}
\usepackage{subcaption}
\usepackage{booktabs} 
\usepackage{amssymb}
\usepackage{amsmath}
\usepackage{mathtools}
\usepackage{booktabs} 
\usepackage{makecell} 
\usepackage[table,xcdraw]{xcolor}

\usepackage{multirow}
\usepackage{booktabs} 
\usepackage{arydshln}  
\newcommand{\sref}[1]{Sec. ~\ref{#1}}
\newcommand{\fref}[1]{Fig. ~\ref{#1}}

\newcommand{\tref}[1]{Table ~\ref{#1}}
\newcommand{\eref}[1]{Eq. ~\ref{#1}}

\usepackage{colortbl}

\newcommand{\revi}[1]{{\color{black}#1}}

\definecolor{lightgray}{gray}{0.9}
\definecolor{mediumgray}{gray}{0.7}
\definecolor{cvprblue}{rgb}{0.21,0.49,0.74}
\usepackage[pagebackref,breaklinks,colorlinks,allcolors=cvprblue]{hyperref}

\def\paperID{*****} 
\def\confName{CVPR}
\def\confYear{2026}

\title{Repurposing 2D Diffusion Models for 3D Shape Completion}

\author{Yao He, Youngjoong Kwon, Tiange Xiang, Wenxiao Cai, Ehsan Adeli\\
Stanford University\\
}

\begin{document}
\twocolumn[{
\renewcommand\twocolumn[1][]{#1}
\maketitle
\vspace{-4em}
\begin{center}
    \centering
    \includegraphics[width=1.0\textwidth]{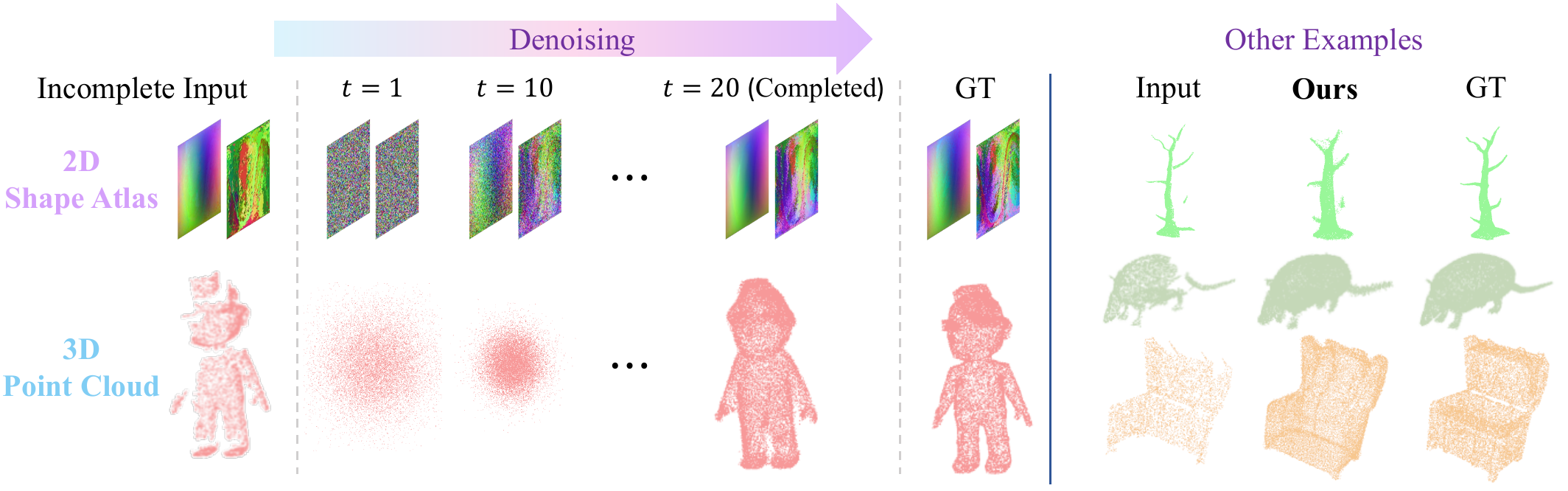}

    \captionof{figure}{We present \textbf{Shape Atlas}, a 2D dense representation of 3D object geometry. \textit{Right}: By serving as a digital twin of 3D assets in the 2D domain, Shape Atlas enables high-quality 3D shape completion using powerful 2D diffusion models. This formulation bridges 2D generative priors and 3D asset generation, allowing shape completion through diffusion in the 2D atlas space and then reconstructed back into full 3D geometry. \textit{Left}: By fine-tuning 2D diffusion models in the atlas domain, our approach generates high-quality 3D point clouds.}
    \label{fig:teaser}
\end{center}
}]
\begin{abstract}
We present a framework that adapts 2D diffusion models for 3D shape completion from incomplete point clouds. While text-to-image diffusion models have achieved remarkable success with abundant 2D data, 3D diffusion models lag due to the scarcity of high-quality 3D datasets and a persistent modality gap between 3D inputs and 2D latent spaces. To overcome these limitations, we introduce the Shape Atlas, a compact 2D representation of 3D geometry that (1) enables full utilization of the generative power of pretrained 2D diffusion models, and (2) aligns the modalities between the conditional input and output spaces, allowing more effective conditioning. This unified 2D formulation facilitates learning from limited 3D data and produces high-quality, detail-preserving shape completions. We validate the effectiveness of our results on the PCN and ShapeNet-55 datasets. Additionally, we show the downstream application of creating artist-created meshes from our completed point clouds, further demonstrating the practicality of our method.
\end{abstract}

    
\section{Introduction}

Shape completion aims to recover missing geometry from partial or noisy 3D observations. In real-world scenarios, data captured by sensors such as LiDAR or RGBD cameras often suffer from occlusion, limited viewpoints, and measurement noise, leading to incomplete or corrupted point clouds. Reliable shape completion is therefore essential for many downstream applications, including autonomous driving~\cite{zeng2018rt3d}, robotics~\cite{zhu2024point, cai2025spatialbot}, and AR/VR~\cite{fusion4d,holoportation,xiang2023rendering,wild2avatar}.

Early approaches addressed this problem through coarse-to-fine reconstruction pipelines, where a global structure is first predicted and then refined by subsequent modules~\cite{dai2017shape,zhou2022seedformer,anchorformer,svdformer}. However, due to their regression-based formulation, these methods often fail to capture fine details; instead, they produce smooth, averaged reconstructions.

Recent advances in generative modeling, particularly diffusion models~\cite{stablediffusion}, have opened new possibilities for shape completion by enabling data-driven hallucination of unobserved parts. Despite their success, progress in 3D diffusion-based models has been constrained by the scarcity of large-scale, high-quality 3D datasets and the inherent complexity of modeling high-dimensional geometric structures. As a result, generative performance in 3D still lags far behind that of 2D image models.

To overcome this limitation, several recent works have leveraged 2D diffusion models by generating multi-view depth or normal images and fusing them back into a 3D representation~\cite{wei2024pcdreamer}. However, this strategy introduces substantial modality inconsistency between the 3D point cloud input and 2D image-based intermediate representations. The need for additional fusion networks increases complexity and prevents full utilization of the rich generative priors already learned by 2D diffusion models.

In this work, we reformulate the 3D shape completion task as a 2D generative problem. At the core of our method lies the \textbf{Shape Atlas}, a compact 2D representation that maps the 3D geometry of point clouds onto a continuous 2D space. This atlas formulation enables us to fully leverage the generative capabilities of pre-trained 2D diffusion models while maintaining consistent conditioning between the input (incomplete point cloud) and the output (completed point cloud) within the same 2D space.

Specifically, we construct the Shape Atlas by projecting the incomplete 3D point cloud using spherical and planar offsetting \cite{xiang2025repurposing}, encoding both point coordinates and surface normals. We then fine-tune a pre-trained Stable Diffusion model by jointly training a conditional U-Net and a denoising U-Net. The conditional U-Net encodes the incomplete Shape Atlas, and its intermediate features are integrated into the denoising U-Net through cross-attention for guided completion. Evaluations on the PCN~\cite{yuan2018pcn} and ShapeNet-55~\cite{shapenet} benchmarks demonstrate that our framework produces high-quality, detail-preserving, and structurally coherent completions, outperforming existing approaches.

\section{Related Work}
\begin{figure*}[t]
	\begin{center}
  \includegraphics[width=\linewidth]{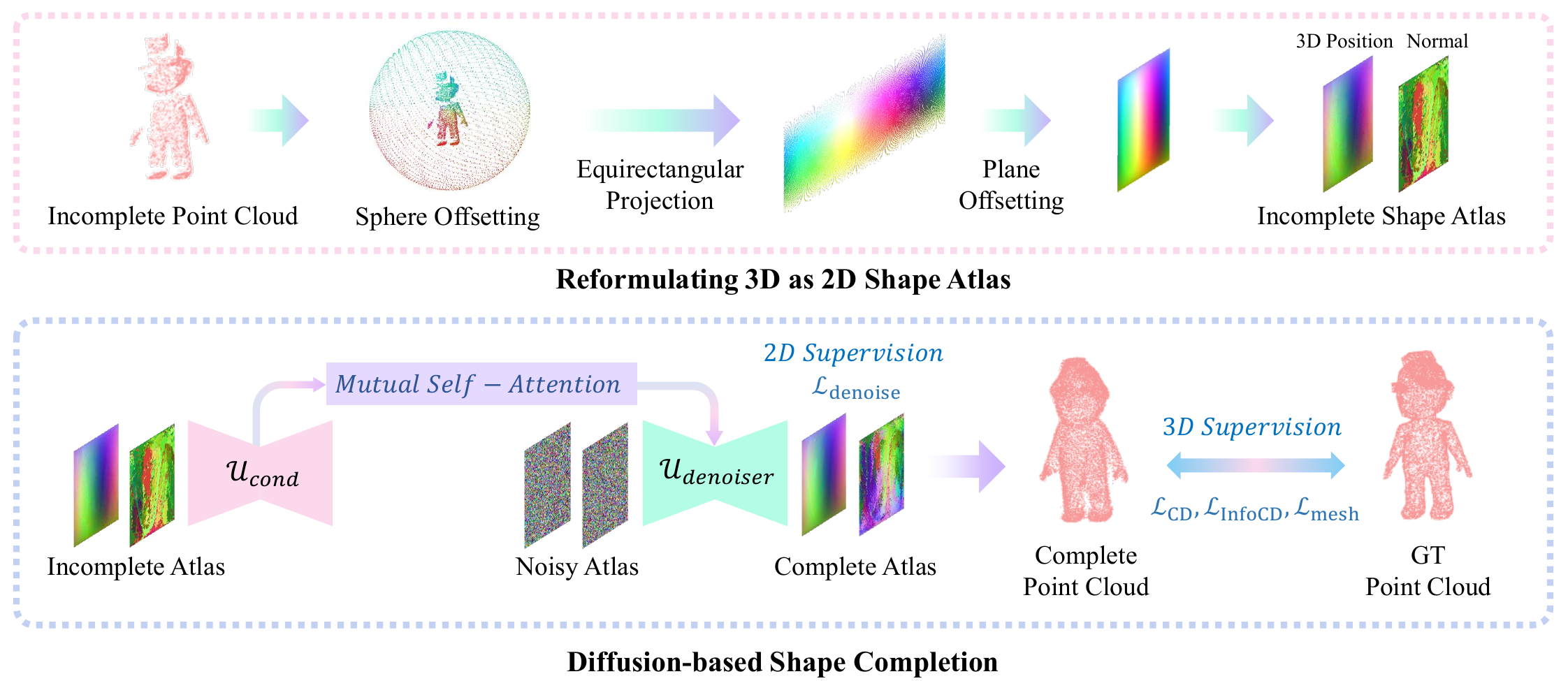}
  \end{center}
  \caption{\textbf{Overview of Our Pipeline.}
\textit{Top}: Representing 3D objects as 2D Shape Atlases. Following \cite{xiang2025repurposing}, the input point cloud is mapped onto the surface of a standard sphere $S$ through \emph{spherical offsetting}, where points sharing the same color indicate the 1-to-1 correspondence. The spherical points are then flattened onto the 2D plane through \emph{plane offsetting}, forming a dense $\sqrt{N}\times\sqrt{N}$ 2D atlas. Our formulation supports both complete and incomplete inputs.
\textit{Bottom}: Training and Inference Pipeline. We train a conditional diffusion model in which 3D reconstruction losses (\textit{i.e.}, $\mathcal{L}_\text{CD}$, $\mathcal{L}_\text{InfoCD}$, $\mathcal{L}_\text{mesh}$) complement the standard 2D diffusion objective (\textit{i.e.}, $\mathcal{L}_\text{denoise}$). The incomplete atlas is processed by a conditioning U-Net $\mathcal{U}_\text{cond}$ to provide control signals to the denoiser $\mathcal{U}_\text{denoiser}$. The denoiser then generates a complete atlas, which is subsequently mapped back to a full 3D point cloud via the inverse of plane offsetting.}
  \label{fig:overview}
\end{figure*}

\subsection{Point Cloud Completion}
The point cloud completion task aims to recover missing geometry from partial or noisy point clouds. Early deep learning approaches~\cite{dai2017shape,han2017high,stutz2018learning,varley2017shape,xie2020grnet} leveraged 3D convolutional networks on voxelized inputs, but these were constrained by high computational and memory costs. PointNet~\cite{qi2017pointnet,qi2017pointnet++} introduced a paradigm shift by directly operating on unordered point sets, enabling more scalable processing. Building on this, PCN~\cite{yuan2018pcn} and subsequent works~\cite{topnet,liu2020morphing,wang2020cascaded,wen2022pmp} adopted a coarse-to-fine folding-based architecture to generate dense point clouds in an end-to-end fashion, conditioned on partial input.
Transformer-based methods~\cite{yu2021pointr,anchorformer,xiang2021snowflakenet,zhou2022seedformer} have since improved completion quality. SnowFlakeNet~\cite{xiang2021snowflakenet} proposes progressively generating child points by splitting their parent points. SeedFormer~\cite{zhou2022seedformer} utilizes upsampling transformers to generate fine details. However, they often suffer from limited contextual understanding, primarily when relying solely on 3D input. To address this, recent methods incorporate image guidance. For instance, MBVN~\cite{hu2019render4completion} employs a conditional GAN to complete missing regions in 2D space and reconstructs 3D point clouds from the completed images. ViPC~\cite{zhang2021view} and Cross-PCC~\cite{wu2023leveraging} use both the partial point cloud and a single-view image for completion. SVDFormer~\cite{zhu2023svdformer} and PCDreamer~\cite{wei2024pcdreamer} take a further step by synthesizing novel views and using them to guide point cloud generation. In particular, SVDFormer leverages self-projected multi-view images to extract 2D features, while PCDreamer employs a diffusion model to generate multi-view depth images. However, due to the limited multi-view consistency of diffusion models, the resulting 3D reconstructions often remain noisy. Further, these approaches require an additional 2D-3D fusion module.
%
%
Importantly, most prior works focus solely on completing the missing points without regard for geometric topology. As a result, meshing these outputs often leads to overly complex and unstructured surfaces. 
Our approach not only refines the point cloud but also explicitly aims to produce artist-created meshes with clean topology, suitable for downstream use in commercial graphics pipelines.

\subsection{2D Representations for 3D Geometry}

Instead of directly learning in 3D space, a growing body of work represents 3D geometry through 2D structures to leverage the advantages of mature 2D learning frameworks.
TriPlane~\cite{eg3d} factorizes a 3D volume into three orthogonal 2D feature planes, enabling 3D generation by interpolating and conditioning on features sampled from these planes.
Following this idea, NFD~\cite{NFD} applies a diffusion model to denoise the triplane representation, which is then decoded into neural fields. CRM~\cite{crm} employs a diffusion model to synthesize multi-view RGB images, while a U-Net generates triplane features conditioned on them. TriplaneGaussian~\cite{triplanegaussian} and DiffGS~\cite{diffgs} further predict triplanes whose Gaussian attributes are directly decoded from the learned features. Pi3D~\cite{pi3d} and HexaGen3D~\cite{hexagen3d} fine-tune diffusion models for triplane generation, but they still rely on additional refinement stages and exhibit limited fidelity.

Another family of approaches adopts pixel-aligned or UV-based 2D representations. Splatter Image~\cite{splatterimage}, PixelSplat~\cite{pixelsplat}, and GPS-Gaussian~\cite{gpsgaussian} predict Gaussian attributes in a pixel-aligned manner, bridging image and 3D domains.
UV-based methods, such as Omages~\cite{omages} and GIMDiffusion~\cite{gimdiffusion}, generate geometry and PBR materials in UV space using pre-trained 2D diffusion models. However, they remain restricted to category-specific generation and depend on handcrafted UV correspondences.

To overcome this limitation, GaussianAtlas~\cite{gaussianatlas} and UVGS~\cite{uvgs} employ spherical mappings to project 3D surfaces onto 2D domains. UVGS introduces multi-layer UV maps to handle occlusions, while GaussianAtlas leverages optimal transport for dense and occlusion-free mappings. Similar to our method, GaussianAtlas fine-tunes a pretrained latent diffusion model (\textit{i.e.}, Stable Diffusion~\cite{stablediffusion}) on 2D representations. However, it is limited to unconditional generation, whereas our approach incorporates a conditional U-Net, also adapted from Stable Diffusion, to enable practical 3D completion from incomplete point clouds.
\section{Method}

\label{sec:method}

\subsection{Motivation}
\label{sec:motivation}

Recent work on repurposing 2D diffusion models for 3D generation has achieved remarkable results in producing high-quality 3D objects from dense 2D representations \cite{xiang2025repurposing, xu2024matters, yang2024atlas, zhang2024gaussiancube, fu2024geowizard}. This paradigm offers several key advantages: it fully leverages recent advances in 2D image and video generation, such as large-scale data and pre-trained models, while simultaneously serving as 2D twins of 3D assets without losing information. For example, it preserves the topology of the original 3D structure, rather than losing depth information entirely as in camera projections.

Existing methods typically assume access to complete observations, and the models are trained directly on perfect data to generate complete geometries~\cite{xiang2025repurposing, zhang2024gaussiancube}. Under this assumption, repurposing a 2D model for 3D generation is straightforward. However, this assumption does not always hold in real-world scenarios, where partial observability is prevalent, such as partial point clouds captured by a single radar or LiDAR scan. Therefore, we propose an approach that leverages 2D priors to tackle the more challenging scenarios on 3D generation from partial observations.

\subsection{Reforumulating 3D as 2D Shape Atlas}
\label{sec:atlas}
In this section, we describe how we transform unordered 3D shape (both complete and incomplete) into dense 2D representations, enabling us to leverage the powerful capabilities of 2D pre-trained generative models. Our pipeline for this transformation is illustrated at the top of \fref{fig:overview}.

\noindent \textbf{Overview} We adopt a strategy similar to recent works that transform 3D shapes into 2D atlases through a two-step process: spherical offsetting and plane offsetting \cite{xiang2025repurposing}. The sphere offsetting begins by hypothesizing a unit sphere centered around the object, populated with $N$ 3D points $\boldsymbol{s} = \{s_i \vert s_i \in \mathbb{R}^3\}$ uniformly distributed on its surface. Each point in the input point cloud $\boldsymbol{p} = \{p_i \vert p_i \in \mathbb{R}^3\}$ is then mapped onto the sphere surface via Optimal Transport (O.T.) \cite{burkard1999linear}. Once positioned, an equirectangular projection is applied to convert the spherical coordinates to flattened 2D coordinates ${m_i \in \mathbb{R}^2}$. In the plane offsetting process, a second O.T. step maps the projected coordinates $m_i$ onto the vertices $n_i \in \mathbb{R}^2$  of a regular $\sqrt{N} \times\sqrt{N}$ 2D grid. As noted in \cite{xiang2025repurposing}, the O.T. step in plane offesting only needs to be computed once, making it inexpensive. However, the O.T. in spherical offsetting depends on the input, which can become computationally expensive, hindering both dataset scalability and inference speed. To address this, we introduce several approaches that reduce the computational complexity.

\noindent \textbf{Improving Sphere Offsetting Efficiency} 
We assume that the input is obtained from an arbitrary scan with an input size of $N_{in}$. Our goal is to map the point cloud onto a fixed spherical points of size $\sqrt{N}\times \sqrt{N}$, which leads to a size mismatch problem. A straightforward solution is to apply O.T. directly to inputs of unequal size. However, since $N$ is large and the computational complexity of O.T. is $\mathcal{O}(n^3)$, this manner is prohibitively expensive. In practice, the input typically comes from a single scan, which is a small subset of the complete point cloud, implying $N_{in} \ll N$. Based on this observation, we first uniformly sample $N_{in}$ points from the complete set of $N$ sphere points, then perform O.T. between the input and this sampled subset. For the remaining points, we apply kNN matching \cite{larose2014k}: each unmatched point is assigned to its nearest matched neighbors, and the same values are propagated. This process also generates a mask that distinguishes points directly matched to the input from those assigned via kNN. This approach drastically reduce the complexity from $\mathcal{O}(N^3)$ to $\mathcal{O}(N^3_{in})$. To further improve efficiency, we leverage the kNN-based Jonker–Volgenant algorithm for linear assignment \cite{chen2020oodanalyzer}, which reduces the O. T. complexity from $\mathcal{O}(n^3)$ to $\mathcal{O}(kn^2)$, where $k$ is the number of nearest neighbors considered as matching candidates.

We encode point clouds with their 3D locations, normals, and masks. This leads to the Shape Atlas shape of $\mathbf{X} \in \mathbb{R}^{\sqrt{N} \times\sqrt{N} \times (\mid \mid \boldsymbol{p} - \boldsymbol{s} \mid \mid + \mid \mid \boldsymbol{n} \mid \mid + \mid \mid \boldsymbol{m} \mid \mid)}$, where $\boldsymbol{p} - \boldsymbol{s}$, $\boldsymbol{n}$  and $\boldsymbol{m}$ denote the offset 3D coordinates, normals and masks, respectively. Furthermore, by optimizing the atlasing process for efficiency, we can scale up the atlas datasets, which is crucial for training a generative model.

\subsection{2D Diffusion for 3D Shape Completion}
\begin{figure*}[t]
	\begin{center}
  \includegraphics[width=1.0\linewidth]{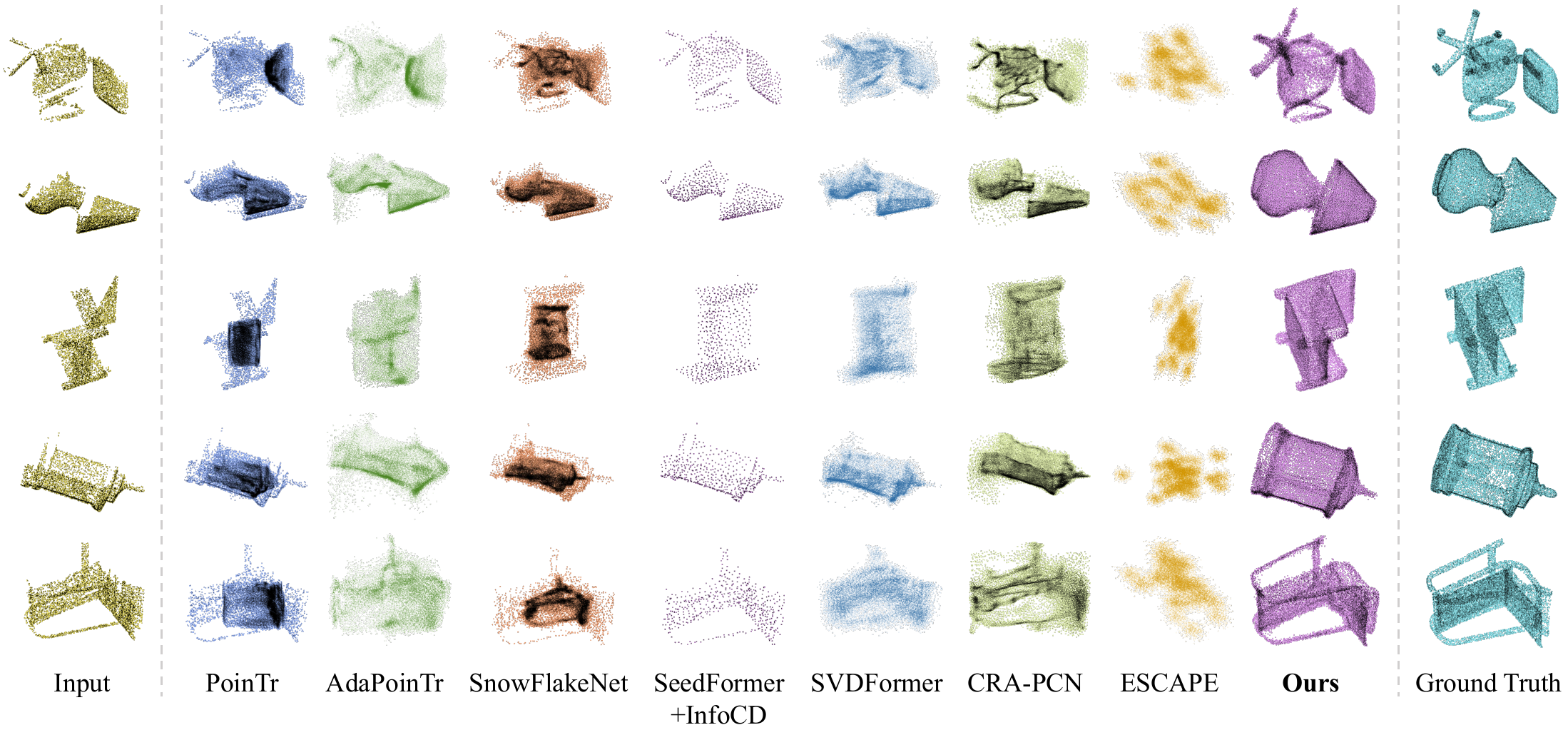}
  \end{center}
  \vspace{-1em}
  \caption{\textbf{Qualitative Comparison of Point Cloud Completion on the PCN Dataset~\cite{yuan2018pcn}.} We show incomplete input point clouds from five representative categories, along with the completed results produced by both baseline methods and our approach. Our approach generates high-quality point clouds while preserving geometric consistency from incomplete inputs.}
  \label{fig:comparison_pcn}
  \vspace{-1.0em}
\end{figure*}

Upon obtaining the Shape Atlases, we fine-tune a latent diffusion model,  \textit{i.e.}, Stable Diffusion~\cite{stablediffusion}, to generate atlases. Following \cite{xiang2025repurposing}, we adapt the diffusion model to operate directly on the atlas without the use of a VAE \cite{pinheiro2021variational}. The training procedure is multi-stage, as described below.

\noindent \textbf{Training Unconditional Diffusion Model}
Our Shape Atlas generation process synthesizes 2D atlases corresponding to complete point clouds, forming a set denoted as $X_c$. Each sample ${\boldsymbol{x}_c} \in {X_c}$ represents a complete Shape Atlas and serves as the clean data $\boldsymbol{x}_0$ in the diffusion process. We fine-tune a denoising U-Net $\mathcal{U}_\text{denoiser}$ on $X_c$ to equip it with the basic capability of generating complete Shape Atlases. Our training strategy follows the velocity-prediction parameterization introduced by Salimans and Ho \textit{et al}.~\cite{ho2022video}. The forward process is defined as:
\begin{equation}
\boldsymbol{x}_t = \alpha_t \boldsymbol{x}_0 + \sigma_t \boldsymbol{\epsilon}, 
\qquad \boldsymbol{\epsilon} \sim \mathcal{N}(\boldsymbol{0},\boldsymbol{I})
\end{equation}
with $\alpha_t=\sqrt{\bar\alpha_t}$, $\sigma_t=\sqrt{1-\bar\alpha_t}$ and $\bar\alpha_t=\prod_{i=0}^{t}(1-\beta_i)$ for a noise schedule ${\beta_{t=0:T}}$. The velocity target is defined as:
\begin{equation}
\boldsymbol{v}_t = \alpha_t\boldsymbol{\epsilon} - \sigma_t\boldsymbol{x}_0
\label{velocity_target}
\end{equation}
At each diffusion step $t$, $\mathcal{U}_\text{denoiser}$ predicts the velocity $\boldsymbol{v}_{t}$, and the model is optimized with the objective
\begin{equation}
\mathcal{L}_\text{denoise}=\mathbb{E}_{\substack{\boldsymbol{x}_0 \sim X_c\\ \epsilon \sim \mathcal{N}(0,I)}}
\left[
\left||
\mathcal{U}_{\text{denoiser}}(\boldsymbol{x}_t, t) - \boldsymbol{v}_t
\right||_2^2
\right]
\label{diffusion_loss}
\end{equation}

\noindent \textbf{Training Conditional Diffusion Model} After fine-tuning the denoising model $\mathcal{U}_\text{denoiser}$ in an unconditional manner, we proceed to train a conditional diffusion model in two stages, where incomplete Shape Atlases $\boldsymbol{x}_i \in X_i$ are the conditioning inputs. Our conditional model follows the ControlNet-style approach \cite{zhang2023adding}. Specifically, conditioning U-Net  $\mathcal{U}_\text{cond}$, which has an architecture identical to $\mathcal{U}_\text{denoiser}$, processes $\boldsymbol{x}_i$ and provides control signals to corresponding layers of the denoising U-Net $\mathcal{U}_\text{denoiser}$ via mutual self-attention \cite{xu2024magicanimate, zhu2024champ}. Finally, $\mathcal{U}_\text{denoiser}$ generates a complete Shape Atlas $\boldsymbol{x}_c$ from pure Gaussian noise. For additional details on architectures, please refer to the supplementary material.

In the first stage, we initialize $\mathcal{U}_\text{cond}$ with the same weights $\boldsymbol{\theta}_0$ as $\mathcal{U}_\text{denoiser}$. We then freeze $\mathcal{U}_\text{denoiser}$ and train $\mathcal{U}_\text{cond}$ by conditioning on $\boldsymbol{x}_i$ to generate $\boldsymbol{x}_c$, using the diffusion loss defined in \eref{diffusion_loss}. This stage yields updated weights $\boldsymbol{\theta}_0^{c}$ for $\mathcal{U}_\text{cond}$. In the second stage of training, we initialize $\mathcal{U}_\text{cond}$ with $\boldsymbol{\theta}_0^{c}$ and $\mathcal{U}_\text{denoiser}$ with $\boldsymbol{\theta}_0$, and jointly train both networks. 

Given that Shape Atlas enables the paradigm of 2D diffusion, it is natural to incorporate additional control, i.e., ControlNet, into the generation process, as is common in classic 2D approaches \cite{zhu2024champ, xu2024magicanimate, zhang2023adding, he2025magicman}. However, extending this idea to 3D is non-trivial, since 2D atlases are constructed to follow a specific topology. A naive application of 2D control methods can hinder the model’s ability to capture high-frequency details. Notably, the velocity prediction allows reconstruction of the initial atlas input via
\begin{equation}
{\boldsymbol{x}}_0 = {\alpha_t \boldsymbol{x}_t - \sigma_t \boldsymbol{v}_t}
\end{equation}
 With plane offsetting, we can inversely map reconstructed input $\hat{\boldsymbol{x}}_0$ to the corresponding pointcloud $\hat{\boldsymbol{p}}_0$. This observation inspires us to guide the training process through additional 3D signals. Therefore, we propose to use a composition of multiple losses during the second training:
\begin{equation}
    \begin{aligned}
    \mathcal{L} =& \lambda_\text{denoise}\mathcal{L}_\text{denoise} + \frac{1}{t+1} \big( \lambda_\text{CD}\mathcal{L}_\text{CD}(\hat{\boldsymbol{p}}_0, \boldsymbol{p}_{gt}) + \\ &\lambda_\text{InfoCD}\mathcal{L}_\text{InfoCD}(\hat{\boldsymbol{p}}_0, \boldsymbol{p}_{gt}) + 
    \lambda_\text{mesh}\mathcal{L}_\text{mesh}(\hat{\boldsymbol{p}}_0, \boldsymbol{M}_{gt})\big)
    \end{aligned}
    \label{losses}
\end{equation}
where $\lambda_\text{denoise}$, $\lambda_\text{CD}$, $\lambda_\text{InfoCD}$ and $\lambda_\text{mesh}$ are tunable parameters. $\mathcal{L}_\text{CD}$ is the chamfer distance (CD) loss, and $\mathcal{L}_\text{mesh}$ is the point-to-mesh-face distance loss where the groud-truth meshes are also provided in our datasets. Minimizing Euclidean distances between matched points as in CD is known to be sensitive to outliers, often leading to clumping artifacts. To address this, we additionally employ InfoCD loss \cite{lin2023infocd} $\mathcal{L}_\text{InfoCD}$, which encourages better point distribution alignment between point clouds and provides a more robust estimation of surface similarity. These losses jointly balance diffusion-based generation with
geometric fidelity.

\section{Dataset}

\label{sec:dataset}
A large-scale dataset with high-quality samples is the de facto foundation for effectively training generative models \cite{gu2023memorization, xu2024matters, kaplan2020scaling, henighan2020scaling}.
Following standard practice for 3D asset generation \cite{wei2024pcdreamer, meshanythingv2, zhu2023svdformer}, we construct our Shape Atlas datasets from ShapeNet-55 \cite{shapenet} and Objaverse \cite{Objaverse}. The source objects are primarily represented in mesh format, i.e., artist-created meshes. We follow the filtering strategy of \cite{meshanythingv2} to retain objects with desired face counts. For each object, we uniformly sample $N$ points on the surface to obtain a complete pointcloud. To generate incomplete point clouds, we sample $N_c$ camera viewpoints around the object and render visibility masks of mesh faces from each view. Points are then selected from the visible faces to form the incomplete pointclouds. Importantly, all the point clouds are transformed into the camera coordinate frame for an ego-centric setup, normalized, and centered at the origin to remove scale and transformation ambiguity. Shape Atlases are then generated for each complete and incomplete point cloud using the method described in \sref{sec:atlas}. We discard undesired samples such as those causing failures in the kNN-based Jonker–Volgenant algorithm. The final dataset consists of roughly 1M paired samples, including complete/incomplete atlases, complete/incomplete point clouds, and ground-truth meshes. We randomly split the datasets into $75\%$ for training, $15\%$ for validation, and $10\%$ for testing. For additional details on datasets and Shape Atlas samples, please refer to
the supplementary material. 

Additionally, the PCN dataset \cite{yuan2018pcn} has been widely adopted as a benchmark for point cloud completion tasks \cite{yuan2018pcn, xie2020grnet, yu2021pointr, wen2022pmp, xiang2021snowflakenet, zhou2022seedformer, chen2023anchorformer, lin2023infocd, zhu2023svdformer, rong2024cra, bekci2025escape}. It is a subset of ShapeNet containing objects from eight categories. We therefore include PCN as an additional benchmark dataset for evaluation. Following the same procedure described above, we generate both partial and complete point clouds, along with their corresponding Shape Atlases. We adhere to the original PCN splits for training, validation, and testing.
\section{Experiments and Results}
\begin{table*}[!htb]
    \centering
    \caption{\textbf{Quantitative Results on the PCN Dataset} ($\ell^{1}$ CD ×$10^{3}$ and F-Score@1$\%$). We show the Chamfer distance for each category while reporting the average of the metrics on the right. The \colorbox{red!25}{red} and \colorbox{yellow!25}{yellow} colors indicate the best and the second best of the methods.}
    \vspace{-2mm}
    \renewcommand{\arraystretch}{1.1}
    \resizebox{0.95\textwidth}{!}{
\begin{tabular}{r|c c c c c c c c|c c}
    \toprule
     Methods & Plane & Cabinet & Car & Chair & Lamp & Couch & Table & Boat & CD-Avg$\downarrow$ & F1$\uparrow$ \\
     \midrule
     PoinTr~\cite{yu2021pointr} & \cellcolor{yellow!25}12.79 & 44.73 & \cellcolor{yellow!25}27.52 & 38.53 & 33.46 & 32.05 & 34.22 & 16.60 & \cellcolor{yellow!25}30.16 & \cellcolor{yellow!25}0.384 \\
     AdaPoinTr~\cite{yu2301adapointr} & 19.23 & 69.78 & 39.56 & 66.80 & 52.66 & 48.50 & 66.08 & 21.33 & 48.59 & 0.260 \\
     SnowFlakeNet~\cite{xiang2021snowflakenet} & 14.10 & 45.44 & 27.54 & 40.09 & 34.90 & 31.40 & \cellcolor{yellow!25}34.02 & 17.33 & 30.82 & 0.327 \\
     \makecell{\footnotesize SeedFormer\\[-2pt] \footnotesize + InfoCD}~\cite{lin2023infocd} & 50.98 & 63.04 & 48.41 & 58.13 & 60.21 & 49.53 & 56.26 & 46.45 & 54.13 & 0.0269 \\
     SVDFormer~\cite{zhu2023svdformer} & 15.84 & 57.53 & 32.69 & 54.13 & 42.05 & 43.33 & 53.20 & \cellcolor{yellow!25}16.31 & 39.86 & 0.302 \\
     CRA-PCN~\cite{rong2024cra} & 21.68 & 77.83 & 37.71 & 77.40 & 59.30 & 53.73 & 75.11 & 21.90 & 54.18 & \revi{0.227} \\
     ESCAPE~\cite{bekci2025escape} & 19.86 & \cellcolor{yellow!25}44.65 & 29.03 & \cellcolor{yellow!25}37.73 & \cellcolor{yellow!25}31.86 & \cellcolor{yellow!25}29.82 & 48.61 & 20.44 & 33.01 & \revi{0.220} \\
     \midrule
     Ours & \cellcolor{red!25}8.86 & \cellcolor{red!25}16.05 & \cellcolor{red!25}13.81 & \cellcolor{red!25}16.03 & \cellcolor{red!25}12.65 & \cellcolor{red!25}16.29 & \cellcolor{red!25}12.58 & \cellcolor{red!25}11.65 & \cellcolor{red!25}13.49 & \cellcolor{red!25}0.480 \\
     \bottomrule
\end{tabular}
    }
    \label{tab:Quantitative_results_PCN}
\end{table*}
\begin{table*}[!htb]
    \centering
    \caption{\textbf{Quantitative Results on the ShapeNet-55 Dataset} ($\ell^{2}$ CD ×$10^{3}$ and F-Score@1$\%$). Chamfer Distance is reported for the categories listed, while the average performance across all categories, including those not listed, is provided in the right-most columns.}
    \vspace{-2mm}
    \renewcommand{\arraystretch}{1.0}
    \resizebox{\textwidth}{!}{
\begin{tabular}{r|c c c c c c c c c c|c c}
    \toprule
    Methods & bench & display & loudspeaker & bookshelf & bathtub & knife & laptop & file & cellular telephone & clock & CD-Avg$\downarrow$ & F1$\uparrow$ \\
    \midrule
    GRNet~\cite{xie2020grnet} & 4.73 & 9.81 & 25.40 & 9.77 & 28.35 & 3.39 & 19.16 & 38.57 & 5.13 & 12.21 & 15.20 & 0.270 \\
    PoinTr~\cite{yu2021pointr} & \cellcolor{yellow!25}1.18 & \cellcolor{yellow!25}5.69 & \cellcolor{yellow!25}19.93 & \cellcolor{yellow!25}2.79 & \cellcolor{yellow!25}19.21 & \cellcolor{yellow!25}0.36 & \cellcolor{yellow!25}11.84 & \cellcolor{yellow!25}33.81 & \cellcolor{yellow!25}3.32 & \cellcolor{yellow!25}10.04 & \cellcolor{yellow!25}10.05 & \cellcolor{red!25}0.455 \\
    \makecell{\footnotesize SeedFormer\\[-2pt] \footnotesize + InfoCD}~\cite{lin2023infocd} & 9.14 & 15.47 & 34.13 & 10.86 & 39.16 & 14.66 & 19.77 & 55.33 & 14.96 & 16.92 & 21.65 & 0.0153 \\
    SVDFormer~\cite{zhu2023svdformer} & 9.42 & 15.34 & 29.92 & 9.14 & 29.90 & 16.98 & 19.98 & 46.76 & 12.29 & 19.19 & 20.25 & 0.0948 \\
    \midrule
    Ours & \cellcolor{red!25}0.61 & \cellcolor{red!25}1.18 & \cellcolor{red!25}1.60 & \cellcolor{red!25}0.94 & \cellcolor{red!25}1.60 & \cellcolor{red!25}0.16 & \cellcolor{red!25}2.12 & \cellcolor{red!25}1.53 & \cellcolor{red!25}0.42 & \cellcolor{red!25}1.81 & \cellcolor{red!25}1.13 & \cellcolor{yellow!25}0.443 \\
    \bottomrule
\end{tabular}
    }
    \label{tab:Quantitative_results_55}
    \vspace{-2mm}
\end{table*}
\label{sec:exp}

In this section, we benchmark and evaluate the effectiveness of our proposed method on the test sets of the datasets described above. We primarily use our samples from PCN~\cite{yuan2018pcn} for benchmarking, as most baseline methods are trained on this dataset. For baselines that are also trained on ShapeNet-55~\cite{shapenet}, we include benchmark on our samples from ShapeNet-55. Ablation studies are conducted on ShapeNet-55 as well.

In addition to the main comparisons on point cloud completion, we demonstrate an additional downstream application. Since artist-created mesh generation methods \cite{meshanything, meshanythingv2, meshtron} require high-quality, complete point clouds to produce clean and well-structured meshes, we use the completed point clouds from our model to generate such meshes. 

\noindent \textbf{Implementation Details}
We set the total number of points per Shape Atlas to $N = 16384 = 128 \times 128$. Each atlas contains 8 channels in total: 3 for point coordinates, 3 for normals, 1 for the visibility mask, and 1 additional dummy channel introduced for training stability. We initialize the 2D U-Net $\mathcal{U}_\text{denoise}$ from Stable Diffusion v1.5 \cite{Rombach_2022_CVPR} by reshaping the input layer from 4 to 8 channels and copying the pre-trained weights accordingly. No text conditioning is used, as we assume the point cloud is the only input, consistent with other baselines for fairness. We employ the AdamW optimizer \cite{loshchilov2017decoupled} with a learning rate of $1\times10^{-5}$ and a batch size of 8 during training. Exponential Moving Average (EMA) and mixed-precision training are used to ensure stable and efficient optimization. We train the unconditional diffusion model for 100K steps. When training the conditional diffusion model, we train for 100k steps during the first stage and 600K steps during the second stages on ShapeNet and Objaverse, followed by fine-tuning on PCN for 320K additional steps. All trainings are conducted on 8 NVIDIA L40S GPUs. During testing, a random view is selected from each test sample for evaluation.

\noindent \textbf{Evaluation Metrics} Following recent work in point cloud completion \cite{wei2024pcdreamer, bekci2025escape}, we use Chamfer Distance under the L1-norm (\textit{i.e.}, CD-L1) as the evaluation metric on the PCN dataset and CD-L2 on ShapeNet-55. For mesh generation quality, we  adopt the following standard metric setting as in \cite{chen2022ndc}: CD-L1, 
Edge Chamfer Distance (ECD), Normal Consistency (NC), Vertex/Face Counts ($\text{\#V}$, $\text{\#F}$), and Vertex/Face Ratios ($\text{V\_Ratio}$, $\text{F\_Ratio}$).

\subsection{Results \& Discussion}

\noindent \textbf{Evaluation on PCN}
On the PCN dataset, we compare our method with two representative approaches in point cloud completion: (1) regression baselines conditioned directly on the partial 3D point clouds-PoinTr~\cite{yu2021pointr}, AdaPoinTr~\cite{yu2301adapointr}, SnowFlakeNet~\cite{xiang2021snowflakenet}, SeedFormer~\cite{zhou2022seedformer}, CRA-PCN~\cite{rong2024cra}, ESCAPE~\cite{bekci2025escape} and
(2) regression baselines conditioned on both 2D inputs (self projected multi view images) and 3D point clouds-SVDFormer~\cite{svdformer}.
As shown in \tref{tab:Quantitative_results_PCN} and \fref{fig:comparison_pcn}, our method outperforms all baselines on all the categories.

\noindent \textbf{Evaluation on ShapeNet-55}
\begin{figure*}[t]
	\begin{center}
  \includegraphics[width=1.0\linewidth]{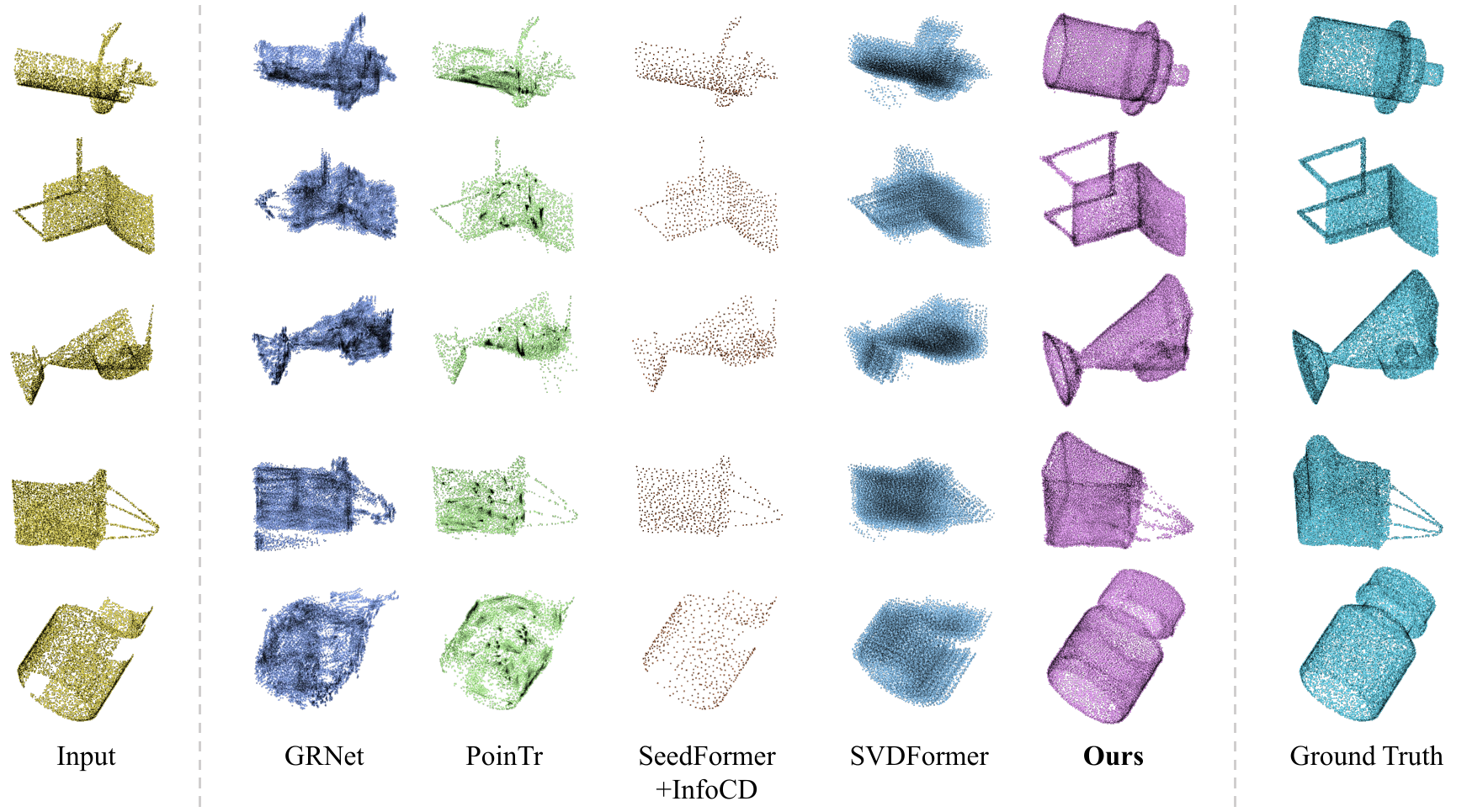}
  \end{center}
  \vspace{-1em}
  \caption{\textbf{Qualitative Comparison of Point Cloud Completion on the ShapeNet-55 Dataset~\cite{shapenet}.} We show incomplete input point clouds from five representative categories, along with the completed results produced by both baseline methods and our approach. Our approach generates high-quality point clouds while preserving geometric consistency from incomplete inputs.}
  \label{fig:comparison_shapenet}
  \vspace{-1.0em}
\end{figure*}

As shown in \tref{tab:Quantitative_results_55} and \fref{fig:comparison_shapenet}, similar to the quantitative results on the PCN benchmark, our method outperforms all baselines in CD across all categories on the ShapeNet-55 dataset and achieves the second best F1 score. The results on ShapeNet together with the results on PCN support our claim that pre-trained 2D diffusion models can be effectively repurposed for 3D shape completion. Notably, our method can also reconstruct fine structures with reasonable accuracy.

\begin{figure}[t]
	\begin{center}
  \includegraphics[width=1\linewidth]{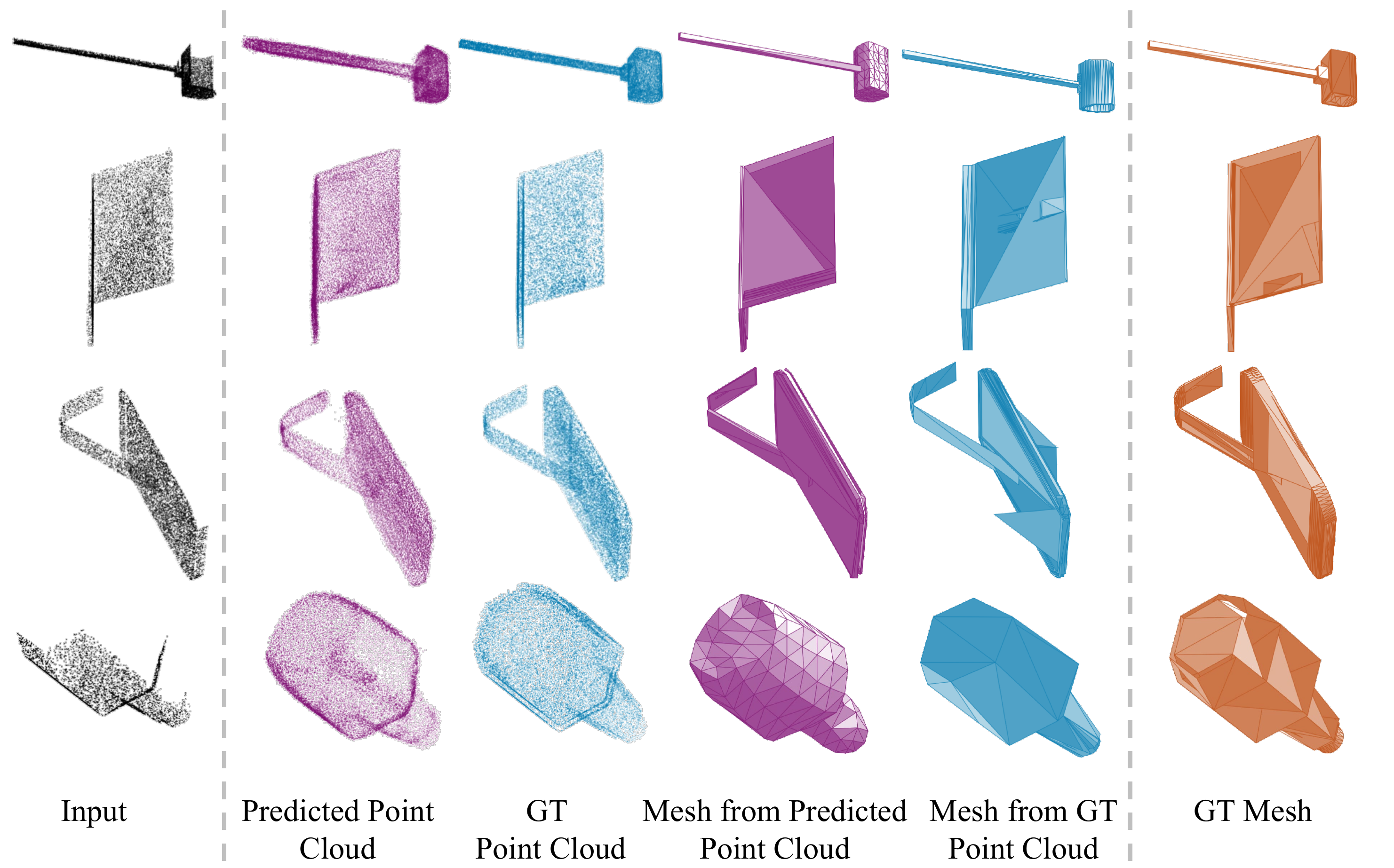}
  	\end{center}
  \caption{\textbf{Qualitative Comparison on Meshes} generated by our predicted point cloud and the ground truth. Meshes generated from our completed point clouds exhibit stable and compact topology, closely matching those produced from the ground truth.}
  \label{fig:mesh}
  \vspace{-0.7em}
\end{figure}

\noindent \textbf{Evaluation on Mesh Generation Quality}
We demonstrate a downstream application of our completed point clouds for artist-created mesh generation on ShapeNet-55. We first apply our method to convert an incomplete partial inputs into a clean and complete point clouds, and then use the off-the-shelf artist-created mesh generation method MeshAnything~\cite{meshanything} to obtain the final meshes. We compare against a competitive scenario in which MeshAnything is directly fed the complete and clean ground-truth point clouds. The results are shown in \tref{tab:mesh} and \fref{fig:mesh}. Using the completed point clouds produced by our approach leads to clean and accurate meshes, and we observe that the resulting meshes preserve stable and compact topology. Given that mesh generation pipelines are highly sensitive to input point cloud quality~\cite{meshtron}, these results validate that our method can generate high-quality point clouds from incomplete inputs for downstream applications.

\begin{table}[t]
  \centering
  \caption{\textbf{Comparison of Artistic-Created Mesh Generation Quality} using (1) our completed point cloud and (2) the ground-truth complete point cloud as inputs to MeshAnything.}
  \label{tab:mesh}
  \resizebox{\linewidth}{!}{
  \begin{tabular}{lccccccc}
    \toprule
    Input & CD-L1$\downarrow$ (×10$^{-2}$) & ECD$\downarrow$ (×10$^{-2}$) & NC$\uparrow$ & \#V$\downarrow$ & \#F$\downarrow$ & V\_Ratio$\downarrow$ & F\_Ratio$\downarrow$  \\
    \midrule
    Ours          & 0.155 & 0.417 & 0.622 & 104.4 & 176.8 & 0.946 & 0.824 \\
    Ground Truth  & 0.161 & 0.405 & 0.631 & 96.62 & 167.9 & 0.870 & 0.786 \\
    \bottomrule
  \end{tabular}
  }
\end{table}
\begin{table}[t]
    \centering
    \caption{\textbf{Quantitative Comparison under Different View Settings} on the ShapeNet-55 dataset ($\ell^{2}$ CD ×$10^{3}$ and F-Score@1$\%$). }
    \vspace{-2mm}
    \renewcommand{\arraystretch}{1.1}
    \resizebox{\linewidth}{!}{
    \begin{tabular}{r|cc|cc|cc}
        \toprule
        \multirow{2}{*}{Methods} 
        & \multicolumn{2}{c|}{View-1} 
        & \multicolumn{2}{c|}{View-2} 
        & \multicolumn{2}{c}{View-3} \\
        \cmidrule(lr){2-3}\cmidrule(lr){4-5}\cmidrule(lr){6-7}
        & CD-Avg$\downarrow$ & F1$\uparrow$ & CD-Avg$\downarrow$ & F1$\uparrow$ & CD-Avg$\downarrow$ & F1$\uparrow$ \\
        \midrule
        GRNet~\cite{xie2020grnet} & 15.20 & 0.270 & 42.45 & 0.0880 & 40.53 & 0.0908 \\
        PoinTr~\cite{yu2021pointr} & \cellcolor{yellow!25}10.05 & \cellcolor{red!25}0.455 & \cellcolor{yellow!25}10.39 & \cellcolor{red!25}0.454 & \cellcolor{yellow!25}10.18 & \cellcolor{red!25}0.456 \\
        \makecell{\footnotesize SeedFormer\\[-2pt] \footnotesize + InfoCD}~\cite{lin2023infocd} & 21.65 & 0.0153 & 29.32 & 0.0105 & 61.99 & 0.0039 \\
        SVDFormer~\cite{zhu2023svdformer} & 20.25 & 0.0948 & 20.81 & 0.0957 & 20.58 & 0.0950 \\
        \midrule
        Ours & \cellcolor{red!25}1.13 & \cellcolor{yellow!25}0.443 & \cellcolor{red!25}1.22 & \cellcolor{yellow!25}0.445 & \cellcolor{red!25}1.05 & \cellcolor{yellow!25}0.442 \\
        \bottomrule
    \end{tabular}
    }
    \label{tab:view_diff}
    \vspace{-2mm}
\end{table}

\subsection{Ablation Studies}
\noindent \textbf{View Consistancy}
\begin{figure}[t]
	\begin{center}
  \includegraphics[width=1\linewidth]{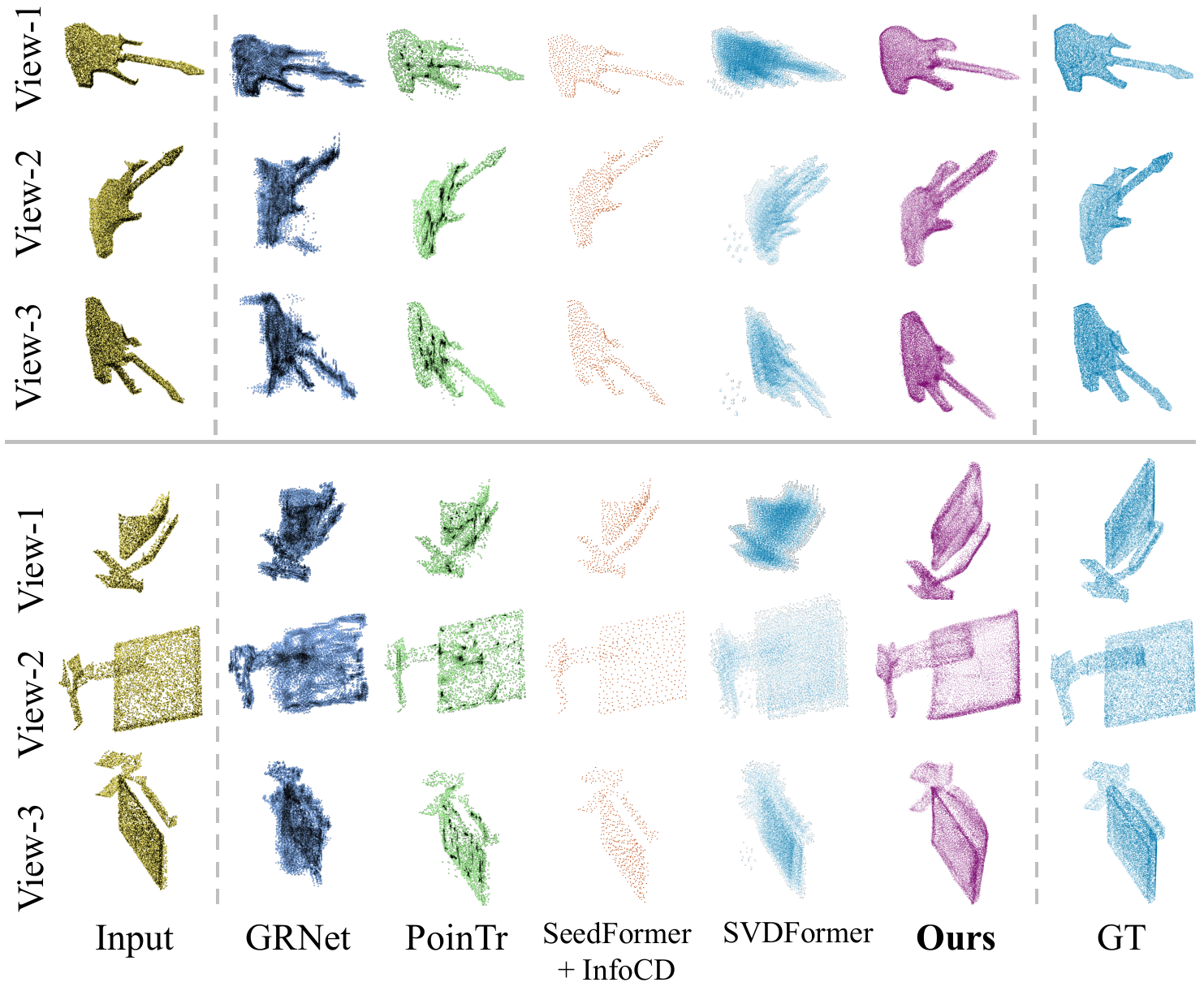}
  	\end{center}
  \caption{\textbf{Qualitative Comparison of Generation Quality of the Same Objects Under Different Views.} Our method consistently generates high-quality point clouds under different views.}
  \label{fig:view_diff}
\end{figure}
View inconsistency is a critical issue in recent point cloud completion methods. As reported in \cite{bekci2025escape}, methods such as \cite{xiang2021snowflakenet, zhou2022seedformer, yu2021pointr, yu2301adapointr, anchorformer} rely on canonical shape representations and therefore suffer significant performance degradation when the input is rotated due to viewpoint changes. Similar degradation can be observed in our results in \tref{tab:Quantitative_results_PCN} and \tref{tab:Quantitative_results_55} when compared to the performance reported in their original papers. To verify that our model and training setup are not affected by canonical representations and are thus view-consistent, we conduct an additional evaluation on ShapeNet-55 using alternative viewpoints from each test sample. The results, shown in \tref{tab:view_diff} and \fref{fig:view_diff}, demonstrate that our approach consistently achieves the best CD-L2 performance and the second-best F1 score, further confirming the effectiveness of our method under viewpoint changes.

\begin{figure}[t]
	\begin{center}
  \includegraphics[width=1\linewidth]{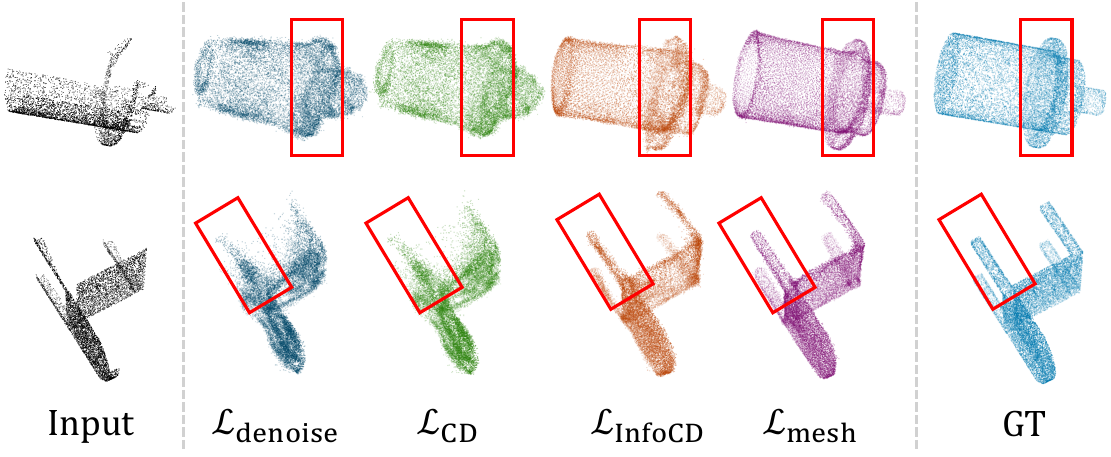}
  	\end{center}
\vspace{-0.5em}
  \caption{\textbf{Qualitative Comparison of The Effect of Different Losses.} Each column corresponds to the addition of a new loss. Fine structures are highlighted to illustrate the improvements.}
  \label{fig:loss_effect}
\end{figure}

\noindent \textbf{Effective of 3D Losses} We adopt several 3D losses to facilitate the training of the diffusion model. In \tref{tab:ablation_loss} and \fref{fig:loss_effect}, we demonstrate the effectiveness of these 3D losses in improving generation quality. The results show that relying solely on the standard diffusion loss and CD produces coarse shapes but fails to capture high-frequency geometric details. Incorporating InfoCD substantially enhances the model’s ability to recover fine structures, and further adding the mesh loss leads to high-quality point cloud completion.

\begin{table}[t]
    \centering
    \caption{\textbf{Ablation Study on the Effect of each Loss Term} on the ShapeNet-55 dataset ($\ell^{2}$ CD ×$10^{3}$ and F-Score@1$\%$). 
    Each row adds one additional loss component to the training objective.}
    \vspace{-2mm}
    \renewcommand{\arraystretch}{1.1}
    \resizebox{\linewidth}{!}{
    \begin{tabular}{p{4.0cm}|>{\centering\arraybackslash}p{2.5cm} >{\centering\arraybackslash}p{2.5cm}}
        \toprule
        {Loss used} & {CD-Avg$\downarrow$} & {F1$\uparrow$} \\
        \midrule
        $\mathcal{L}_{\text{denoise}}$ & 3.65 & 0.252 \\
        $\mathcal{L}_{\text{denoise}}$, $\mathcal{L}_{\text{CD}}$ & 3.05 & 0.273 \\
        $\mathcal{L}_{\text{denoise}}$, $\mathcal{L}_{\text{CD}}$, $\mathcal{L}_{\text{InfoCD}}$ & 1.75 & 0.392 \\
        $\mathcal{L}_{\text{denoise}}$, $\mathcal{L}_{\text{CD}}$, $\mathcal{L}_{\text{InfoCD}}$, $\mathcal{L}_{\text{mesh}}$ & 1.13 & 0.443 \\
        \bottomrule
    \end{tabular}
    }
    \label{tab:ablation_loss}
    \vspace{-2mm}
\end{table}

\noindent \textbf{2D Pre-Trained Model Serves as A Better Initialization}
\begin{figure}[t]
	\begin{center}
  \includegraphics[width=1\linewidth]{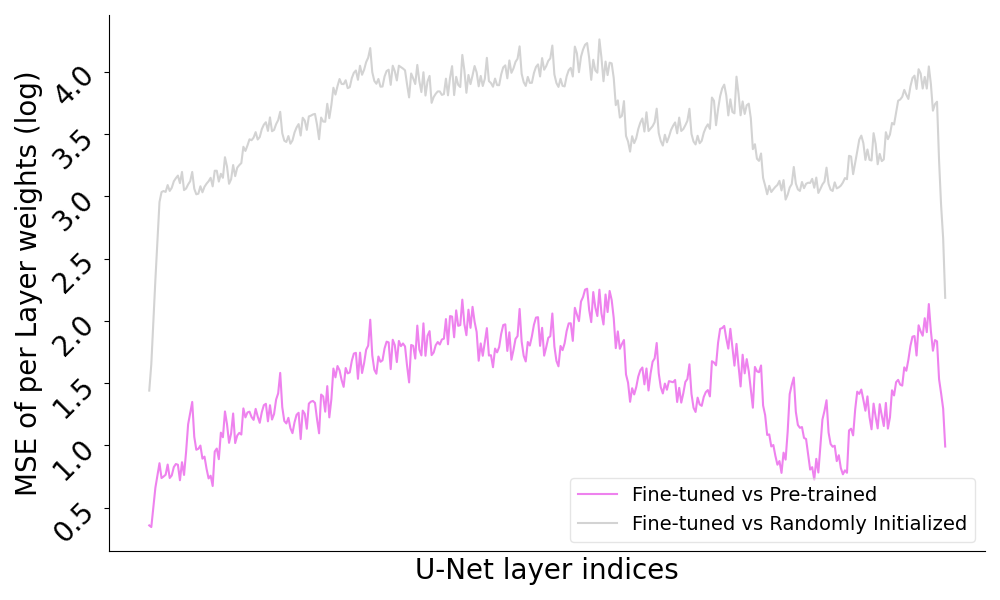}
  	\end{center}
  \caption{\textbf{Log Scale Smoothed MSE of per-Layer
Weights between Different U-Nets}. Finetuning from a pre-trained LD U-Net leads to smaller weight differences.}
  \label{fig:weight difference}
  \vspace{-0.5em}
\end{figure}
In \fref{fig:weight difference}, we plot the log scale mean squared errors (MSE) of the per-layer weight differences between our finetuned U-Net and two baselines: a randomly initialized U-Net and the pre-trained latent diffusion (LD) U-Net. We observe that the weights change only slightly when fine-tuning from the pre-trained LD U-Net, while the differences relative to a randomly initialized U-Net are significantly larger. Even for the layer with the largest update, the deviation from the LD initialization remains far smaller than that from random initialization. These findings indicate that the LD U-Net offers a superior starting point, enabling the model to converge more effectively for 3D shape completion.

\section{Conclusions}
We present an effective shape completion approach from partial observations such as incomplete point clouds. At the core of our approach is a novel formulation that expresses a 3D shape as a 2D atlas, which enables finetuning of a pretrained 2D diffusion model and allows us to draw on the abundant knowledge contained in large scale 2D latent diffusion models. This formulation also addresses the modality gap between the condition and the output by unifying them in the same 2D atlas space. Experiments demonstrate that our method achieves state of the art performance in 3D point cloud completion. We further provide downstream results in artist created mesh generation, which demonstrate the practicality of our method.

{
    \small
    \bibliographystyle{ieeenat_fullname}
    \bibliography{main}
}
\clearpage
\setcounter{page}{1}
\maketitlesupplementary

\appendix

\section{Implementation Details}
\subsection{Reformulating 3D as 2D Shape Atlas}
In \sref{sec:atlas}, two kNN matching steps are involved. For the kNN matching of unassigned points, we set $k=1$, whereas for the kNN configuration used in the Jonker–Volgenant algorithm, we set $k=50$. Overall, generating a Shape Atlas takes less than 30 seconds for a point cloud of size 16,384, and less than 10 seconds for incomplete point clouds with fewer than 10,000 points. We further employ extensive parallelization to accelerate this process during dataset construction.
\subsection{Model Architecture}
Both the denoising U-Net $\mathcal{U}_\text{denoiser}$ and the conditioning U-Net $\mathcal{U}_\text{cond}$ are implemented using the \texttt{UNet2DConditionModel} from the Hugging Face Diffusers library\footnote{\url{https://huggingface.co/docs/diffusers}}, initialized with the Stable Diffusion v1.5 U-Net weights provided by Hugging Face. The reference controller for the U-Net is implemented following MagicAnimate \cite{xu2024magicanimate}, where corresponding transformer blocks in $\mathcal{U}_\text{denoiser}$ and $\mathcal{U}_\text{cond}$ share a common feature bank. During inference, transformer blocks from $\mathcal{U}_\text{cond}$ write features into the bank, while transformer blocks from $\mathcal{U}_\text{denoiser}$ concatenate their hidden-state outputs with the stored features from the bank. The concatenated features are then passed through a self-attention layer followed by a cross-attention layer, providing the conditioned features needed by the denoising U-Net.
\subsection{Details on Training the Conditional Diffusion Model}
We adopt a curriculum training strategy. In the first stage, we train for 100K steps using only the diffusion loss $\mathcal{L}_{\text{denoise}}$. In the second stage, we again train for 100K steps with only the diffusion loss, after which we add the Chamfer Distance loss $\mathcal{L}_{\text{CD}}$ for the next 100K steps with a weighting of $\lambda_{\text{CD}} = 0.05$. However, we observe that incorporating $\mathcal{L}_{\text{CD}}$ yields only minimal improvements, so we introduce the InfoCD loss $\mathcal{L}_{\text{InfoCD}}$ in the following 100K steps, setting $\lambda_{\text{CD}} = 0.00$ and $\lambda_{\text{InfoCD}} = 0.2$. Adding the InfoCD loss drastically improves generation quality. In the remaining 400K steps, we further incorporate the mesh loss $\mathcal{L}_{\text{mesh}}$ with a weighting of $\lambda_{\text{mesh}} = 100$ to better capture high-frequency geometric details. The full training on ShapeNet and Objaverse takes approximately 11 days. We then fine-tune on PCN for an additional 320K steps using the same loss weights, which requires roughly 5 days to complete.

\section{Details on Datasets}
\begin{figure*}[t]
	\begin{center}
  \includegraphics[width=\linewidth]{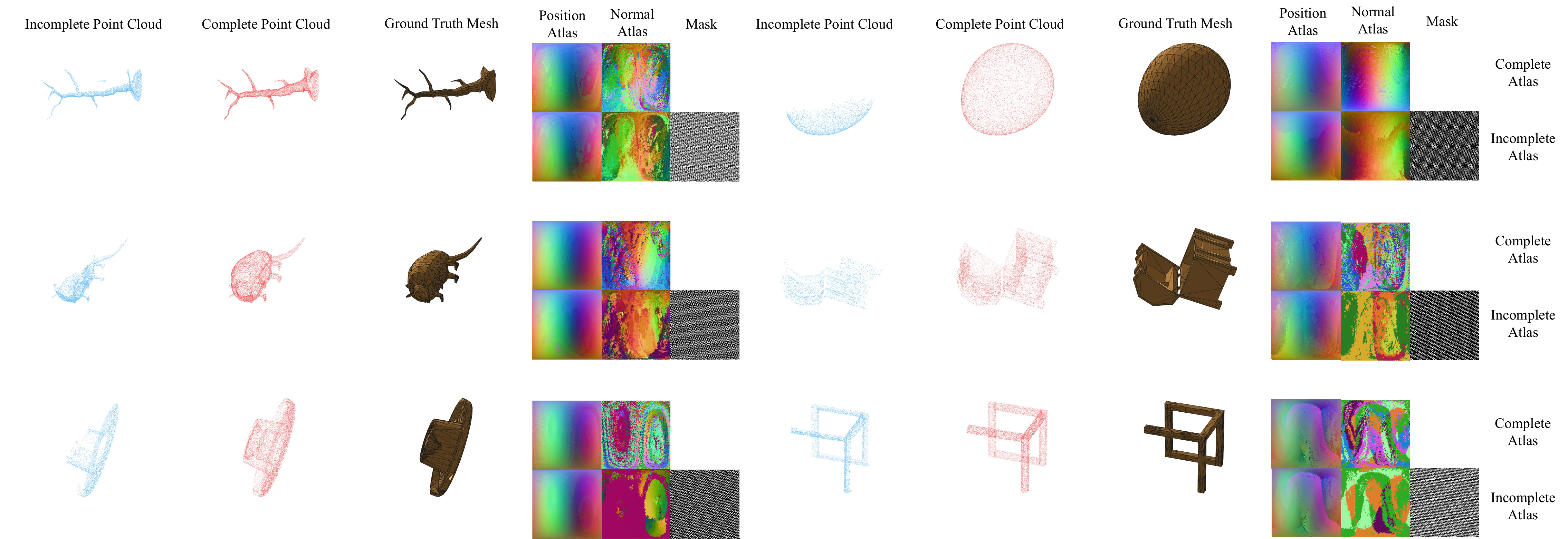}
  \end{center}
  \caption{\textbf{Data Examples.} We present six representative data samples from our processed Objaverse dataset, including incomplete point clouds, complete point clouds, meshes, and their corresponding Shape Atlases. Our processed ShapeNet-55 and PCN datasets follow the same data format. Each sample exhibits complex geometric structures. For clearer visualization of the visible and missing regions in the incomplete point clouds, we manually rotate each example. Overall, our dataset encompasses a large number of diverse and geometrically rich samples.}
  \label{fig:data-samples}
\end{figure*}

\begin{figure*}[t]
	\begin{center}
  \includegraphics[width=\linewidth]{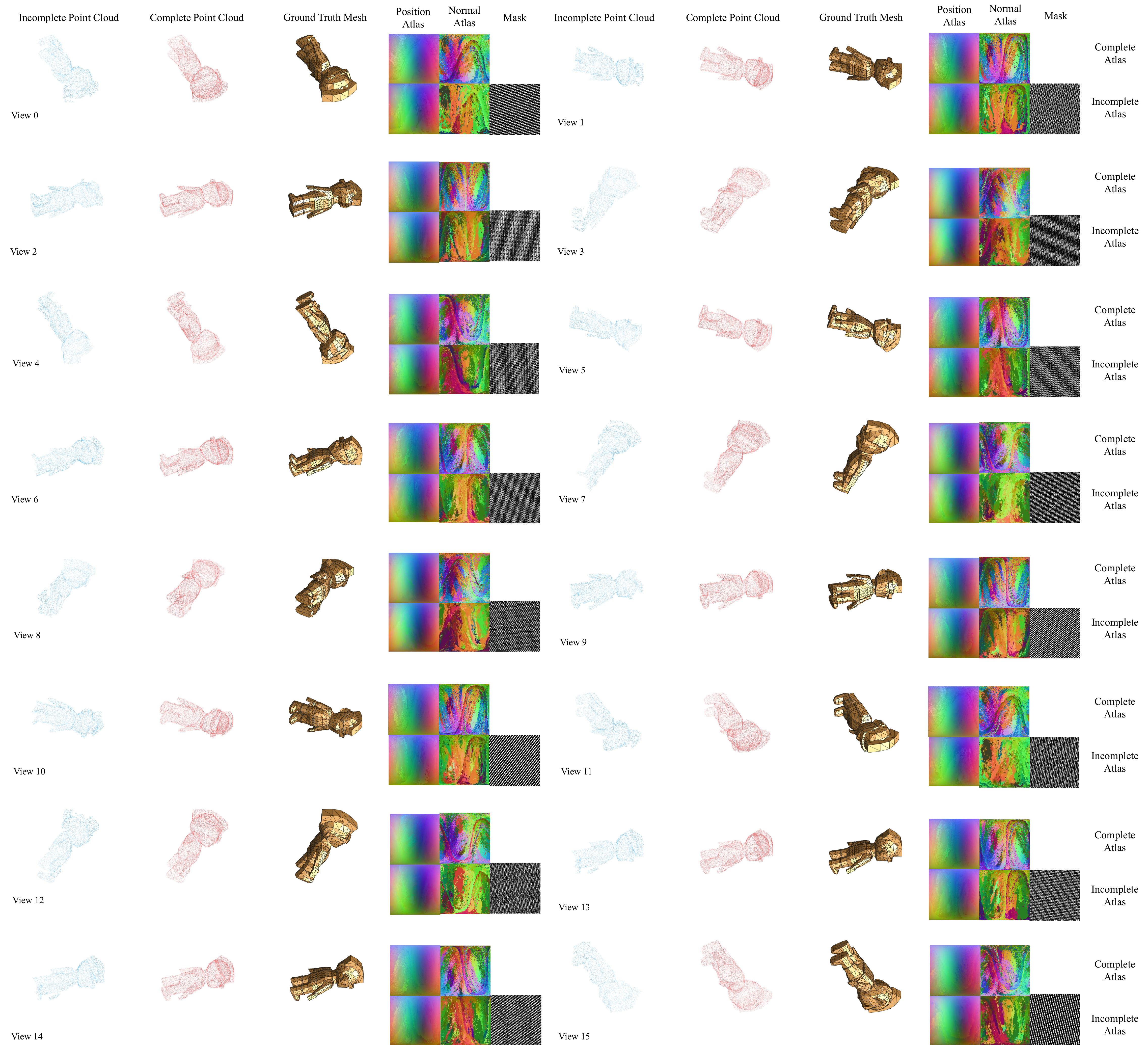}
  \end{center}
  \caption{\textbf{Data Samples from the Same Object under Different Views.} We present all data samples obtained from the same object across multiple viewpoints. Each object is shown in its original orientation to enable clear comparison across views.}
  \label{fig:data-views}
\end{figure*}

For ShapeNet-55 and Objaverse, we filter out objects with fewer than 1600 mesh faces to construct our dataset, while for PCN we directly use the full dataset without filtering. We uniformly sample 16 camera viewpoints around each object, where each camera is oriented toward the object center with an added random offset in the range of \([-0.1,\, 0.1]\). Face visibility masks are rendered using \texttt{Nvdiffrast}\footnote{\url{https://github.com/NVlabs/nvdiffrast}}
, with cameras configured at a resolution of $512\times512$ pixels and a field of view of $120^{\circ}$. For each pair of complete and incomplete point clouds and the corresponding ground-truth mesh from the same object, we first transform all data into the camera coordinate frame. We then set the centroid of the incomplete point cloud as the origin and shift the point clouds and the mesh accordingly. This normalization removes translation ambiguity and ensures consistent alignment for all subsequent processing. Our dataset includes 10538 objects from ShapeNet-55 and 60401 objects from Objaverse. \fref{fig:data-samples} and \fref{fig:data-views} provide several visualization of data samples.

\section{More Results on Objaverse}
\begin{figure*}[t]
	\begin{center}
  \includegraphics[width=1\linewidth]{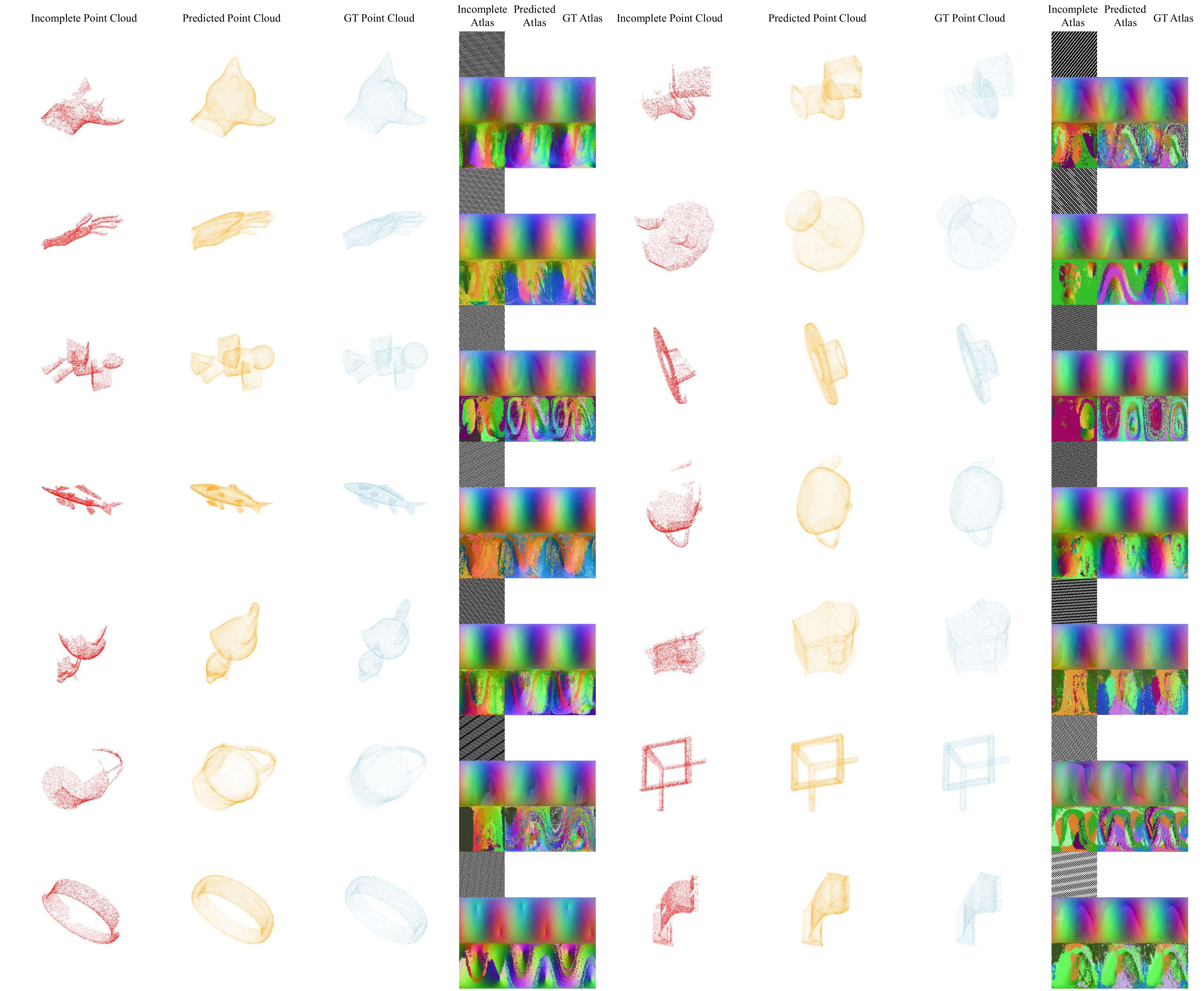}
  	\end{center}
  \caption{\textbf{More Qualitative Results on Diverse Objaverse Objects.} We present generated point clouds for a wide variety of objects from Objaverse, demonstrating the broad generative capability of our model.}
  \label{fig:more_result}
  \vspace{-0.7em}
\end{figure*}

\begin{figure}[t]
	\begin{center}
  \includegraphics[width=1\linewidth]{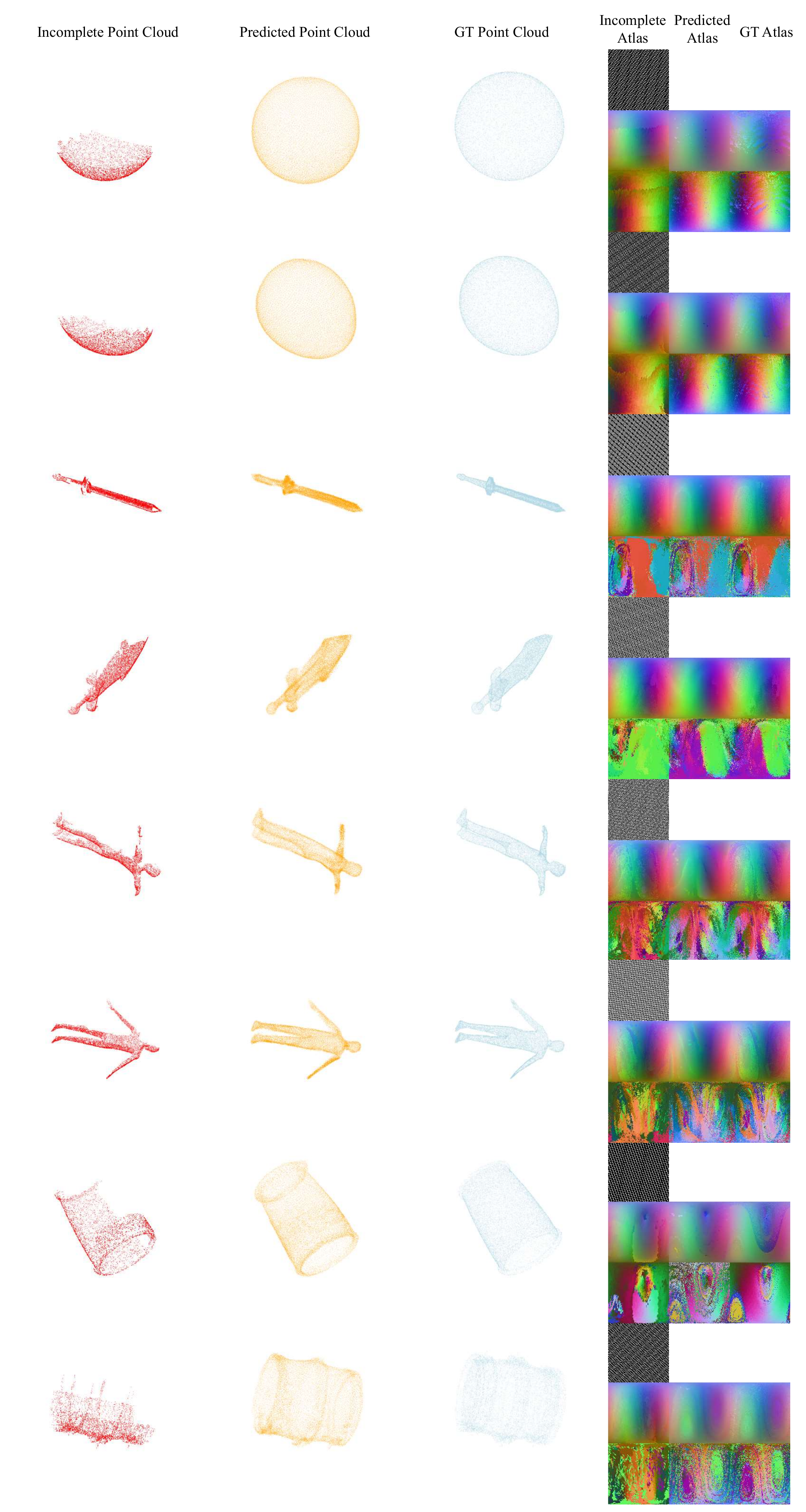}
  	\end{center}
  \caption{\textbf{More results on visually similar objects from Objaverse.} We show generated point clouds of visually similar objects, grouped every two rows, to demonstrate the capability of our model to capture fine geometric details and diverse variations while maintaining structural consistency across different instances.}
  \label{fig:more_result_2}
  \vspace{-0.7em}
\end{figure}

We present additional results on our test set from Objaverse. Compared to ShapeNet-55 and PCN, which exhibit relatively limited structural diversity, Objaverse contains a much broader set of objects with more complex and finely detailed geometry. We show qualitative results from Objaverse in \fref{fig:more_result} and \fref{fig:more_result_2}, demonstrating our model’s capability to generate diverse outputs while maintaining consistency across different instances of objects within similar categories. \tref{tab:Quantitative_results_Objaverse} presents the quantitative results on the Objaverse dataset. Since existing baselines do not train or evaluate on this dataset to the best of our knowledge, we report our measurements along with the direct inference results of PoinTr \cite{yu2021pointr}. Additionally, we introduce Objaverse as an evaluation dataset for point cloud completion for the first time, providing a significantly more diverse and challenging testing scenarios. We hope that this new paradigm on Objaverse will establish a new benchmark for future work in point cloud completion and broader 3D vision research.

\begin{table}[!htb]
    \centering
    \caption{\textbf{Quantitative Results on the Objaverse Dataset} ($\ell^{2}$ CD ×$10^{3}$ and F-Score@1$\%$). We report only the averaged metrics for our method.}
    \vspace{-2mm}
    \renewcommand{\arraystretch}{1.1}

\resizebox{0.9\linewidth}{!}{%
\begin{tabular}{p{4.0cm}|>{\centering\arraybackslash}p{2.5cm} >{\centering\arraybackslash}p{2.5cm}}
    \toprule
     Methods & CD-Avg$\downarrow$ & F1$\uparrow$ \\
     \midrule
     PoinTr \cite{yu2021pointr} & 79.04 & 0.387 \\
     Ours & 1.69 & 0.472 \\
     \bottomrule
\end{tabular}
}

    \label{tab:Quantitative_results_Objaverse}
\end{table}


\section{Limitations}
Our work has several limitations that highlight promising directions for future research. First, we do not incorporate text conditioning to control the generation process. Integrating text-guided diffusion could enable more flexible, semantically aligned, and user-controllable 3D shape completion. Second, our training dataset remains small compared to contemporary large-scale 2D image and video diffusion datasets, which limits the diversity and generalization capacity of the model. Although we introduce multiple 3D reconstruction losses to improve the recovery of high-frequency structures, the model still struggles to fully capture fine geometric details. Exploring richer geometric, perceptual, or topology-aware losses may further enhance fidelity in these challenging regions. Finally, due to computational constraints, our training uses a batch size of only 8. Future work may benefit from larger batch sizes and increased compute, which could lead to improved performance, stability, and overall generative quality.


\end{document}